\documentclass[letterpaper, 10 pt, conference]{ieeeconf}  % Comment this line out if you need a4paper
\usepackage[acronym]{glossaries}
\usepackage{bm}
\usepackage{graphicx} 
\usepackage{svg}
\usepackage{subcaption}
\usepackage{xcolor}
\usepackage{hyperref}
\usepackage{tabularx}
\usepackage[font=footnotesize]{caption}
\usepackage[font=footnotesize]{subcaption}
\definecolor{posgreen}{RGB}{26,150,65}
\definecolor{negred}{RGB}{165,0,38}
\definecolor{neblue}{RGB}{100,149,237}
\usepackage{pgfplots}
\usepgfplotslibrary{fillbetween}
\usepgfplotslibrary{statistics}
\usepackage{booktabs}  % For professional lines like \toprule and \midrule
\usepackage{multirow}  % For the \multirow command in the Category column
\usepackage{amsmath}   % For mathematical notation like \times
\makeglossaries

\newacronym{jssp}{JSSP}{job-shop scheduling problem}
\newacronym{agv}{AGV}{automatic guided vehicle}
\newacronym{jsspt}{JSSPT}{job-shop scheduling problem with transportation resources}
\newacronym{drl}{DRL}{deep reinforcement learning}
\newacronym{rl}{RL}{reinforcement learning}
\newacronym{pdr}{PDR}{priority dispatching rule}
\newacronym{dr}{DR}{dispatching rule}
\newacronym{gnn}{GNN}{graph neural network}
\newacronym{marl}{MARL}{multi-agent reinforcement learning}
\newacronym{fifo}{FIFO}{first-in-first-out}
\newacronym{gin}{GIN}{graph isomorphism network}
\newacronym{spt}{SPT}{shortest processing time}
\newacronym{smpt}{SMPT}{shortest machine processing time}
\newacronym{lpt}{LPT}{longest processing time}
\newacronym{mwr}{MWR}{most work remaining}
\newacronym{lwr}{LWR}{least work remaining}
\newacronym{fdd/mwr}{FDD/MWR}{flow due date per most work remaining}
\newacronym{mor}{MOR}{most operations remaining}
\newacronym{lor}{LOR}{least operations remaining}
\newacronym{fcfs}{FCFS}{first come first serve}
\newacronym{sput}{SPUT}{shortest pick-up time}
\newacronym{scta}{SCTA}{shortest completion time of in-transport tasks}
\newacronym{scpt}{SCPT}{shortest completion time of in-transport tasks with pick-up time}
\newacronym{ga}{GA}{genetic algorithm}
\newacronym{ml}{ML}{machine learning}

\newacronym{ppo}{PPO}{proximal policy optimization}
\newacronym{ctde}{CTDE}{centralised training with decentralised execution}
\newacronym{mappo}{MAPPO}{multi-agent PPO}
\newacronym{mdp}{MDP}{Markov decision process}
\newacronym{wr}{WR}{win rate}
\newacronym{rpi}{RPI}{relative percentage increase}
\newacronym{mlp}{MLP}{multi-layer perceptron}
\newacronym{nn}{NN}{neural network}
\newacronym{ci}{CI}{confidence interval}
\newacronym{eput}{EPUT}{earliest pick-up time}
\newacronym{est}{EST}{earliest start time}
\newacronym{ert}{ERT}{earliest ready time}
\newacronym{eat}{EAT}{earliest arrival time}
\newacronym{eft}{EFT}{earliest finish time}
\newacronym{tts}{TTS}{travel time to start}
\newacronym{lb}{LB}{lower bound}
\newacronym{ols}{OLS}{ordinary least squares}
\newacronym{vif}{VIF}{variance inflation factor}
\newacronym{e2e}{E2E}{end-to-end}

\IEEEoverridecommandlockouts                              % This command is only needed if 
\usepackage{pgfplots}
\usepgfplotslibrary{fillbetween}
\title{\LARGE \bf
An Analysis of the Coordination Gap between Joint and Modular Learning for Job Shop Scheduling with Transportation Resources
}

\author{Moritz Link$^{1}$, Jonathan Hoss$^{1}$, and  Noah Klarmann$^{1}$% <-this % stops a space
\thanks{*Supported by the Chips Joint Undertaking and its members, 
including top-up funding by National Authorities, within the Cynergy4MIE project (Grant Agreement No. 101140226).}
\thanks{$^{1}$Faculty of Management and Engineering, 
University of Applied Sciences, 83024 Rosenheim, Germany,
        Correspondence: {\tt\small moritz.link@th-rosenheim.de}}%
\thanks{This paper has been accepted for presentation at the IEEE 22st International Conference on Automation Science and Engineering (CASE 2026)}
}

\begin{document}

\maketitle
\thispagestyle{empty}
\pagestyle{empty}

%%%%%%%%%%%%%%%%%%%%%%%%%%%%%%%%%%%%%%%%%%%%%%%%%%%%%%%%%%%%%%%%%%%%%%%%%%%%%%%%
\begin{abstract}
Efficient job-shop scheduling with transportation resources is critical for high-performance manufacturing. 
With the rise of "decentralized factories", multi-agent reinforcement learning has emerged as a promising 
approach for the combined scheduling of production and transportation tasks.
Prior work has largely focused on developing novel cooperative architectures while overlooking the question of when joint training is necessary. 
Joint training denotes the simultaneous training of job and automatic guided vehicle scheduling agents, 
whereas modular training involves independently training each agent followed by post-hoc integration. 
In this study, we systematically investigate the conditions under which joint training is essential for optimal performance in the job-shop scheduling problem with transportation resources.
Through a rigorous sensitivity analysis of resource scarcity and temporal 
dominance, we quantify the coordination gap --- the performance difference between these two training modalities. 
In our evaluation, joint training outperforms the majority of dispatching rule combinations and modular training approaches.
However, the coordination gap advantage diminishes in bottleneck environments, 
particularly under severe transport and processing constraints. 
These findings indicate that modular training represents a viable alternative in environments where a single scheduling task dominates. 
Overall, our work provides practical guidance for selecting between training modalities based on environmental conditions, enabling decision-makers to optimize reinforcement learning-based scheduling performance.
\end{abstract}

%%%%%%%%%%%%%%%%%%%%%%%%%%%%%%%%%%%%%%%%%%%%%%%%%%%%%%%%%%%%%%%%%%%%%%%%%%%%%%%%
\section{INTRODUCTION}
 
In today's competitive manufacturing environment, optimal scheduling plays a key role in enhancing flexibility, productivity, cost-efficiency, and resource management.
With the integration of \glspl{agv} for material handling, intralogistics scheduling has gained significant relevance. 
This problem, often defined as \gls{jsspt}, extends the classical \gls{jssp} by incorporating transportation 
constraints \cite{berterottiereFlexibleJobshopScheduling2024, nouriClassificationSchemaJob2016}. 
Consequently, the simultaneous optimization of production operations and vehicle-based transportation is critical for 
the development of future high-performance manufacturing environments \cite{blazewiczSchedulingTasksVehicles1991}.

The $NP$-hard \gls{jssp} is commonly addressed using exact methods, heuristics, and metaheuristics, each with different trade-offs in solution quality and computational cost. While exact methods are often impractical for real-time use, \glspl{dr} offer fast but sub-optimal decisions, and \glspl{ga} achieve higher solution quality but are computationally intensive, limiting their applicability in real-time scheduling applications \cite{bagheriArtificialImmuneAlgorithm2010, luoDynamicSchedulingFlexible2020}.

To overcome these limitations, researchers have increasingly applied \gls{drl} to scheduling variations, leveraging its adaptability and generalization capabilities in complex decision-making tasks \cite{yuanResearchFlexibleJob2025}. 
To address the distinct requirements of transportation control and operation scheduling, research has shifted from single-agent \gls{rl} toward \gls{marl}, focusing on cooperative decentralized architectures \cite{liRealTimeSchedulingFlexible2025, chengCooperativeAgentDeep2025,popperSimultaneousProductionAGV2021a, zhouModularCoordinatedMultiagent2024}.

Despite its potential, \gls{marl} is frequently hindered by non-stationarity, increased state-space complexity, partial observability, and the multi-agent credit 
assignment problem. 
These challenges result in degraded training stability and inconsistent policy convergence \cite{bahrpeymaReviewApplicationsMultiagent2022}. While specialized 
training frameworks have been proposed to coordinate agent updates \cite{chengCooperativeAgentDeep2025,zhouModularCoordinatedMultiagent2024}, 
they typically assume centralized control over all agents and the environment. This assumption is often unrealistic in modern "decentralized factories", 
where subsystems are integrated from multiple vendors. Furthermore, centralized training can be computationally expensive when individual agents or components are replaced. 
Finally, the incorporation of shared information, such as embeddings, creates interdependencies that restrict the system's modularity and 
flexibility in dynamic production settings \cite{bahrpeymaReviewApplicationsMultiagent2022, leitaoAgentbasedDistributedManufacturing2009}.

Given the shift toward multi-agent architectures, this study investigates the conditions under which joint training is essential for \gls{jsspt} versus when modular training is sufficient. This research evaluates \gls{drl} performance across two modalities: a joint setting, characterized by the concurrent training of job and \gls{agv} scheduling agents to foster learned cooperation through \gls{marl}, and a modular setting, where agents are trained independently and integrated post-hoc. While prior \gls{rl} research for \gls{jsspt} has focused on proposing cooperative architectures, the question of when joint training is necessary remains largely unexplored. Unlike framework-centric studies, we provide a rigorous sensitivity analysis of the coordination gap, quantifying performance differentials between both training modalities. By evaluating performance across varying resource intensities --- the relative scarcity of transportation resources and temporal-dominance ratios, representing the balance between transportation and processing durations ---we identify the specific conditions under which the performance of independent agents diverges significantly from their coordinated counterparts. 
The main contributions of this study are summarized as follows:
\begin{itemize}
    \item Introduction of a novel training framework that enables independent agent learning while maintaining compatibility for joint evaluation.
    \item A systematic sensitivity analysis of joint versus modular \gls{rl} training for \gls{jsspt}, identifying conditions under which joint training is essential to effectively coordinate both schedulers.
    \item Characterization of the non-linear impact of resource scarcity and temporal-dominance on the coordination gap, elucidating the functional interdependence of logistics and production scheduling.
\end{itemize}

\section{RELATED WORK}

In recent years, several studies have addressed the integrated problem of job scheduling and \gls{agv} scheduling. 
Table \ref{tab:agv_job_scheduling_comparison} provides an overview of \gls{rl}–based approaches to this problem.
For both the job scheduler and the \gls{agv} scheduler, the table defines the scheduling policy type as well as the corresponding action space. The policy type is categorized as either rule-based, where decisions are made according to predefined rules, or \gls{rl}-based, where a learned policy directly determines scheduling actions.
The action type can be either rule-based, where the policy selects a predefined rule, or \gls{e2e}, where scheduling entities such as operations, jobs, or \glspl{agv} are selected directly. While some approaches formulate both job scheduling and \gls{agv} scheduling as learning tasks, others represent the \gls{agv} scheduler using rules and optimize only the job scheduler component through \gls{rl}. 
Furthermore, many studies employ joint training, in which both policies are optimized simultaneously.

However, the impact of different training paradigms has received limited attention. In particular, it remains unclear whether jointly training both scheduling policies leads to superior performance compared to training the job scheduler and \gls{agv} scheduler separately.

\begin{table}[b]
\vspace{-0.5cm}
\caption{Literature comparison of \gls{agv} and job scheduling approaches ( \acrfull{e2e}, operations ($O$), jobs ($J$)), and action space ($A$)).}
\label{tab:agv_job_scheduling_comparison}
\setlength{\tabcolsep}{4pt} 
\small
\begin{tabular}{ccccc}
\toprule
\textbf{paper} &
\textbf{\gls{agv} type} &
\textbf{\gls{agv} $A$} &
\textbf{job type} &
\textbf{job $A$} \\
\midrule

\cite{yuanResearchFlexibleJob2025}
& rule-based
& \textemdash
& \gls{rl}
& \gls{e2e}: $O$\\

\cite{liRealTimeSchedulingFlexible2025}
& \gls{rl}
& rule-based
& \gls{rl}
& rule-based\\

\cite{chengCooperativeAgentDeep2025}
& rule-based
& \textemdash
& \gls{rl}
& \gls{e2e}: $O$\\

\cite{popperSimultaneousProductionAGV2021a}
& \gls{rl}
& \gls{e2e}: $O$
& \gls{rl}
& \gls{e2e}: $O$\\

\cite{zhouModularCoordinatedMultiagent2024}
& \gls{rl}
& \gls{e2e}: \gls{agv}
& \gls{rl}
& \gls{e2e}: $O$\\

\cite{wangDynamicIntegratedScheduling2024}
& \gls{rl}
& rule-based
& \gls{rl}
& rule-based\\

\cite{shenTransformerbasedMultiagentReinforcement2026}
& \gls{rl}
& \gls{e2e}: \gls{agv}
& \gls{rl}
& \gls{e2e}: $J$\\

\cite{wangMultiAgentDeepReinforcement2026}
& rule-based
& \textemdash
& \gls{rl}
& \gls{e2e}: $O$\\

\cite{zhangIntegratedOptimizationMethod2026}
& \gls{rl}
& \gls{e2e}: \gls{agv}
& \gls{rl}
& \gls{e2e}: $O$\\

\cite{yangHGAMPPOUnifiedHeterogeneous2026}
& \gls{rl}
& \gls{e2e}: \gls{agv}
& \gls{rl}
& \gls{e2e}: $O$\\

\bottomrule

Ours (joint)
& \gls{rl}
& \gls{e2e}: \gls{agv}
& \gls{rl}
& \gls{e2e}: $O$\\

Ours (job)
& rule-based
& \textemdash
& \gls{rl}
& \gls{e2e}: $O$ \\

Ours (\gls{agv})
& \gls{rl}
& \gls{e2e}: \gls{agv}
& rule-based
& \textemdash\\
\bottomrule

\end{tabular}
%\vspace{-0.5cm}
\end{table}

\section{METHODOLOGY}

\subsection{Problem Formulation}
Consider a set of $n$ independent jobs, $\mathcal{J} = \{J_{1}, J_{2}, \dots, J_{n}\}$, and a set of $m+2$ machines, 
$\mathcal{M} = \{M_{l},M_{u}, M_{1}, \dots, M_{m}\}$, where $M_{l}$ and $M_{u}$ are the load and unload machines from which jobs enter and leave the system. 
Each machine contains an input and an output buffer to store arriving and processed operations. 
A job $j \in \mathcal{J}$ is associated with a fixed, job-specific routing that defines an ordered sequence of 
$m+1$ operations $O_{j,i}, 1 \leq i \leq m +1$. Each operation must be processed on a specific machine $\mu_{j,i} \in \mathcal{M}$ 
for a processing time $p_{j,i} > 0$. 
The last operation $O_{j,m+1}$ represents the release to the unload machine $M_{u}$ with $p_{j,m+1} = 0$. 
Additionally, \glspl{agv} are required to transport operations between machines. 
Specifically, an \gls{agv} picks up an operation from a machine's output buffer and 
delivers it to the subsequent machine's input buffer for further processing. 
A set $V = \{V_{1}, V_{2}, \dots, V_{k}\}$ comprises $k$ \glspl{agv}, with potentially asymmetric transportation times $t(M_{i}, M_{j})$ 
between machines $M_{i}$ and $M_{j}$. 
The input buffers are infinite and maintain the order in which operations are assigned to a machine, without capacity restrictions. 
Conversely, the output buffers provide infinite capacity without an ordering constraint, enabling an \gls{agv} to pick up completed operations in any order.
At the beginning, all \glspl{agv} start at the load machine. The goal of the \gls{jsspt} is to minimize the makespan 
$\min C_{\max}$, with $C_{\max} = \max_{j \in \mathcal{J}} c_{j,m+1}$, where $c_{j,m+1}$ denotes the completion time of job $j$ at the unload machine. 
The completion time is defined as $c_{j,i} = s_{j,i} + p_{j,i}$ with the operation's start time $s_{j,i}$. 
The transport completion time $c_{T_{j,i}}$ 
is determined by the start time $s_{T_{j,i}}$ and 
the travel time $t(\cdot)$ between machine locations $\mu_{j,i-1}$ and $\mu_{j,i}$ as $c_{T_{j,i}} = s_{T_{j,i}} + t(\mu_{j,i-1}, \mu_{j,i})$. 
To solve the \gls{jsspt}, the conditions for the \gls{jssp} are enhanced by transportation conditions. 
For the classic \gls{jssp}, the operations have to be processed in sequential order, expressed 
as $O_{j,1} \xrightarrow[]{} O_{j,2} \xrightarrow[]{} \dots \xrightarrow[]{}O_{j,m+1}$. 
Additionally, machines can process one operation at a time without preemption. 
The \glspl{agv} are restricted by unit-capacity, with no preemption permitted during transportation. 
An operation can only be transported after its predecessor completes processing on its machine: $s_{T_{j,i}} \geq c_{j,i-1}$, 
and the processing of the operation starts after transportation: $s_{j,i} \geq c_{T_{j,i}}$. 
In addition, $s_{T_{j,i}}$ depends on the pending tasks of the \gls{agv}:

\begin{equation}
    s_{T_{j,i}} \geq \max(c_{j,i-1}, \text{idle}_u + t(\text{loc}_u, \mu_{j,i-1})),
\end{equation} where $\text{idle}_u$ denotes the time \gls{agv} $u$ 
finishes its pending tasks, and $\text{loc}_u$ the machine location where $u$ becomes idle. 

\subsection{Markov Decision Process Formulation}
We define the \gls{mdp} by the tuple $(N, S, A, P, R, \gamma)$. $N$ defines the number of agents, $S$ is the finite state space, 
and $A = \prod_{i=1}^{n} A_i$ is the joint action space with $A_i$ denoting the agent $i$'s action space. $P$ represents the transition 
probability $P(s_{t+1}| s_{t}, \bm{a})$ to state $s_{t+1}$ from state $s_{t}$ with joint action $\bm{a} \in A$. $R$ returns a 
scalar reward for a given state $s$ and $\bm{a}$. $\gamma \in [0, 1]$ is the discount factor. In our definition, $N = 2$ in the multi-agent setting or $N = 1$ for the single-agent training. 

\textbf{State:} 
The global state $S_t$ is composed of two local states $S_{t}^{o}$ and $S_{t}^{agv}$, representing the states for the job scheduler 
and \gls{agv} scheduler at timestep $t$. $S_{t}^{o}$ is a disjunctive graph, denoted as $G = \{V,E\}$, with vertices $V$ and edges $E$. Vertices $V = \{O,\mathcal{M}\}$ comprise operations $O$ and machines $\mathcal{M}$. 
Edges are divided into disjunctive edges $D$, encoding precedence constraints between operations, and conjunctive edges $C$, representing machine–operation assignments. 
In this formulation, all operations within a job are connected sequentially, and each operation is bidirectionally connected to its processing machine. This graph representation enables multichannel communication across machine resources while preserving the job precedence flow. The feature set for a vertex $v \in V$ depends on its type, yet all vertices share a common base representation:
1) a binary indicator $S(v)$, which is defined as 1 only if $v$ is already scheduled, and 
2) a binary indicator $T(v)$, which takes the value 1 only if $v$ is a machine vertex. 
In addition, a vertex $v_o \in O$ for operation $O_{j,i}$ contains a \gls{lb} scalar defined as:
\begin{equation}
\text{\gls{lb}}_{j,i} = \max_{%
  \substack{
    s \in \text{scheduled } O_j \\
    s < i \text{ for job } j
  }%
} 
c_{j,s} + \sum_{%
    \substack{
        k \in \text{ unscheduled } O_j\\
        k \leq i}^{} 
    }%
p_{j,k}, 
\end{equation}
and a machine vertex $v_m \in \mathcal{M}$ incorporates the ratio of scheduled to total operations for this machine:
\begin{equation}
 r_{v_m} = \frac{n_{sched}(v_m)}{n},
\end{equation} where $n$ is the number of jobs and $n_{sched}(v_m)$ the number of scheduled operations assigned to machine $v_m$. 
These distinct features denote the type-specific completion ratio. Finally, the $\text{\gls{lb}}_{j,i}$ is normalized based on all other vertices in $O$.

The state $S_{t}^{agv}$ is defined by a set of features for each \gls{agv} $u$. These features comprise six scalars:
\gls{eput}, \gls{est}, \gls{ert}, \gls{tts}, \gls{eat}, and \gls{eft}.
They are conditioned upon the specific operation $O_{j,i+1}$ selected by the job scheduling agent. 
For this operation, the start machine $M_{s}$ describes the location of the preceding operation $O_{j,i}$ and $M_{t}$ the target machine as $O_{j,i+1}$'s processing machine. A transport task comprises the traveling from the \gls{agv}'s location to $M_{s}$ and 
the transport to $M_{t}$. $\text{\gls{eput}}_j$, with $\text{\gls{eput}}_j = c_{j,i}$, is the completion time of $O_{j,i}$ and provides the earliest possible pick-up time for $u$. 
The feature $\text{\gls{est}}_m$ returns the next available time point for $M_{t}$, 
representing the earliest moment the machine can commence processing the next assigned operation $O_{j,i+1}$. It is defined as the maximum completion time $c_{p,l}$ among the set $\mathcal{S}_{M_t}$ of operations $O_{p,l}$ currently scheduled on $M_t$,
$\text{\gls{est}}_m = \max_{O_{p,l} \in \mathcal{S}_{M_t}} \{ c_{p,l} \}$. 
$\text{\gls{ert}}_u$ is the next idle time point of $u$, representing the completion of all pending tasks. 
$\text{\gls{tts}}_u$ is the travel duration (empty transport) from $u$'s location to $M_{s}$, $\text{\gls{tts}}_u = t(\text{loc},M_{s})$, 
with loc denoting the machine where $u$ will be available for the selected task. 
$\text{\gls{eat}}_u$ is the earliest possible arrival time of $u$ at $M_{s}$. 
By combining the time to the idle location and the travel time from there 
to $M_{s}$, $\text{\gls{eat}}_u = \text{\gls{ert}}_u+\text{\gls{tts}}_u$. 
Finally, $\text{\gls{eft}}_u$ represents the earliest finish time of the total transport task. 
It combines $\text{\gls{eat}}_u$ with the transportation duration $t(M_{s}, M_{t})$ as $\text{\gls{eft}}_u=\text{\gls{eat}}_u + t(M_{s}, M_{t})$.
All features are scaled between $[0,1]$. 
For $\text{\gls{eat}}_u$, $\text{\gls{ert}}_u$, $\text{\gls{tts}}_u$, and $\text{\gls{eft}}_u$, scaling is performed relative to the corresponding values of the other \glspl{agv}. The quantities $\text{\gls{eput}}_j$ and $\text{\gls{est}}_m$ are scaled with respect to their related in-processing items: $\text{\gls{eput}}_j$ relative to the ready times of other processing operations, and $\text{\gls{est}}_m$ relative to other processing machines.

\textbf{Actions:} $A = A_{o} \times A_{agv}$, where $A_{o}$ is the set of unfinished operations that can be scheduled at a decision time 
step $t$ and $a_{o} \in A_{o}$ represents the selected operation. $A_{agv}$ is the set of compatible \glspl{agv} for the operation $a_{o}$. 
The final \gls{agv} action is $a_{agv} \in A_{agv}$. Combined they form the joint action $\bm{a} = a_{o} \times a_{agv}$. 

\textbf{Transition:} With the joint action $\bm{a_{t}}$, the environment transitions from $s_t$ to a new state $s_{t+1}$. 
A new state does not represent a discrete forward movement in physical time but denotes the next valid state that fulfills the \gls{jsspt} conditions 
to schedule another operation-\gls{agv} pair after including $\bm{a_{t}}$. The transition to $s_{t+1}$ is deterministic with $P(s_{t+1}| s_{t}, \bm{a_{t}}) = 1$ and occurs immediately 
after the joint action $\bm{a_t}$ is applied.

\textbf{Reward:} The optimization target is to minimize the makespan $ C_{\max}$. 
Consequently, the reward signal is formulated as the negative makespan:
\begin{equation}
R =
\begin{cases}
-\dfrac{ C_{\max}}{\text{\gls{lb}} \cdot s}, & \text{if the schedule is complete} \\
0 & \text{otherwise,}
\end{cases}
\end{equation}
where $s$ is a static scaling factor and \gls{lb} represents the lower bound over all jobs at the beginning of the scheduling process with transportation times. 
While sparse, this reward structure ensures that the agents optimize for the global objective ($C_{max}$) rather than local processing efficiencies, 
avoiding greedy but sub-optimal scheduling behaviors. 
To prevent incentivizing the agents to select infinitely invalid decisions, action masking is applied to invalid operations.

\textbf{Policy:} 
Policy $\pi_{\theta} (\bm{a_{t}}|S_t)$ is a combination of the two sub-policies $\pi (a_{o}|S_{t}^{o})$ and $\pi (a_{agv} | S_{t}^{agv},a_{o})$, 
where the action of the \gls{agv} policy depends on the action of the job policy as described in the state definition.

\subsection{Baseline Solvers and Network Structure}
\subsubsection{Baselines}
We use ten \glspl{dr} for the operation selection and four for the \gls{agv} selection, 
known for their efficient performance \cite{liRealTimeSchedulingFlexible2025, leiMultiactionDeepReinforcement2022a, liResearchDynamicJob2025, changSchedulingFlexibleJobShop2024}. 
The baseline solvers are constructed by combining each operation-selection rule with each \gls{agv}-selection rule. 
For the operation selection the following \glspl{dr} are implemented:
\gls{spt}, \gls{smpt}, \gls{lpt}, \gls{mwr}, \gls{lwr}, \gls{fdd/mwr}, \gls{mor}, \gls{lor}, random, and \gls{fcfs}. 
The \gls{agv} rules are: random, \gls{sput}, \gls{scta}, and \gls{scpt}.

\subsubsection{Encoder and Decoder Structure}
The proposed methodology employs a multi-agent approach to address the \gls{jsspt}, where job scheduling and \gls{agv} selection are handled by decoupled agents. 
In this study, the job scheduling agent is a \gls{gnn} and employs the graph representation $G$ from $S_{t}^{o}$. 
As a basic feature encoder, the \gls{gin} \cite{dc993c6e028241629fb9ca557bd0eba9} is utilized and defined as:
\begin{equation}
    h_v^{(l)} = MLP_{\theta_l}^{(l)} \left( (1 + \epsilon^{(l)}) \cdot h_v^{(l-1)} + \sum_{u \in \mathcal{N}(v)} h_u^{(l-1)} \right),
\end{equation} where $h_v^{(l)}$ is the representation of node $v$ at iteration $l$. 
$MLP$ is a \gls{mlp} with parameters $\theta_l $ and $\epsilon$ a learnable parameter. 
$\mathcal{N}(v)$ are the neighboring nodes of $v$. 
After $L$ iterations, a graph embedding $h_G$ is obtained by a readout function and combines all node embeddings. 
We set $L=2$, $\epsilon=0$, and use for each layer $l$ a two-layered $MLP$ with a hidden dimension of 64 and $\text{ReLU}$ activation. 
The global graph representation $h_G$ is derived by computing the mean over all final node embeddings $h_v^{(L)}$.

The job scheduler's decoder is a $MLP$ with three layers and $\tanh$ activation functions. 
The initial two possess an output dimension of 64, followed by a layer that computes a single logit. 
By concatenating each $h_v^{(L)}$ with $h_G$, the input for the decoder comprises local and global information. 
This network is applied to all operation node embeddings to generate the logits used for action selection.

The \gls{agv} scheduling agent is defined as a three-layered $MLP$ with $\tanh$ activation functions. 
The first two have an output dimension of 16 and the last layer computes the logit. 
Applying this network to all \gls{agv} embeddings yields a vector of logits for action sampling.

The critic employs the same \gls{gin} feature encoder. Its decoder leverages the graph embedding $h_G$ to compute a state value via a three-layered \gls{mlp} with $\tanh$ activation functions. 
The initial two layers possess an output dimension of 64, followed by a final layer that outputs the scalar state value.
The disjunctive graph embedding $h_G$ serves as a sufficient proxy for the total system state, as it implicitly captures the 
temporal dynamics and completion times that already account for transportation delays. This approach reduces the dimensionality of the critic's input, 
thereby facilitating more stable value estimation.

\subsection{Training Framework}

To evaluate the training dynamics of joint versus modularly trained agents, 
two distinct training frameworks are implemented within an identical environment to ensure consistent conditions. 
\Gls{ppo}\cite{schulmanProximalPolicyOptimization2017} is selected as the underlying \gls{rl} algorithm, 
following its strong performance in prior literature. 
To differentiate these training modalities, 
their respective influences on the \gls{mdp} formulation are described. 
During joint training, both schedulers are represented by learnable agents, as described previously. 
Therefore, the $\pi (a_{o}|S_{t}^{o})$ is represented by the $GNN$ and $\pi (a_{agv} | S_{t}^{agv},a_{o})$ by the $MLP$. 
The single-agent \gls{ppo} algorithm is enhanced by its multi-agent variation MAPPO \cite{yuSurprisingEffectivenessPPO2022}. 
Under this variation, 
both agents undergo concurrent training with independent parameter updates while being optimized through a joint objective. 

For the modular training, the job and \gls{agv} policy learning are decoupled into two independent training runs. 
However, to satisfy the \gls{mdp} definition, the joint action $\bm{a_{t}}$ has to consist of both scheduler decisions. 
By replacing one policy with a baseline \gls{dr}, two distinct settings are defined: 
\begin{itemize}
  \item Job policy training: $\pi (a_{o}|S_{t}^{o})=GNN$ and $\pi (a_{agv} | S_{t}^{agv},a_{o}) = \text{\gls{dr}}$
  \item \gls{agv} policy training: $\pi (a_{o}|S_{t}^{o}) = \text{\gls{dr}}$ and $\pi (a_{agv} | S_{t}^{agv},a_{o}) = MLP$
\end{itemize}
A modular solver is constructed by pairing a job scheduling agent and an \gls{agv} scheduling agent, 
with the training \glspl{dr} as solver identifiers. 
$S_c$ defines the set of concatenated solvers with $S_c = \{ J_i\_A_j \mid i \in \{1, \dots, 4\}, j \in \{1, \dots, 10\} \}$ and the index denotes the 
training \gls{dr} as described in Table~\ref{table_modalities_solvers}. 
The \glspl{dr} distinguish the concatenated solver combinations. Our joint solver is identified as J0\_A0.
\begin{table}[t]
\vspace{0.2cm}
\caption{Training modalities and solvers}
\vspace{-0.2cm}
\label{table_modalities_solvers}
\begin{center}
\begin{tabular}{cccc}
\toprule
\textbf{modality} &  \textbf{id} $\bm{i,j}$  & $\bm{Ji}$ \textbf{\gls{dr}}&  $\bm{Aj}$  \textbf{\gls{dr}}\\ 

\midrule
joint  & 0 & - & - \\
\midrule
\multirow{7}{*}{modular} 
     & 1 & \gls{scta} & \gls{fdd/mwr} \\
     & 2 & \gls{scpt} & \gls{mor} \\
     & 3 & random & \gls{lwr} \\
     & 4 & \gls{sput} & \gls{spt} \\
     & 5 & - & \gls{mwr} \\
     & 6 & - & random \\
     & 7 & - & \gls{fcfs} \\
     & 8 & - & \gls{smpt} \\
     & 9 & - & \gls{lor} \\
     & 10 & - & \gls{lpt} \\

\bottomrule
\end{tabular}
\end{center}
\vspace{-0.5cm}
\end{table}

\section{EXPERIMENTS AND RESULTS}
\subsection{Performance Metrics}
To evaluate the performance of the solvers, this work utilizes the \gls{rpi} \cite{chengCooperativeAgentDeep2025} and the \gls{wr}. The \gls{wr} is used to select between \gls{dr} combinations. For the solver of interest $i$ and a baseline solver $y$, the \gls{rpi} is defined as:
\begin{equation} \label{eq:rpi}
    RPI_i = - \frac{C_{\max}^i-C_{\max}^{y}}{C_{\max}^{y}}\times 100.
\end{equation}
The \gls{wr} is defined as the average over the independent wins $w(i, y)$:
\begin{equation}
    w(i, y) =
    \begin{cases}
        1, & \text{if } C_{\max}^i < C_{\max}^{y} \\
        0, & \text{otherwise}
    \end{cases}.
\end{equation}

The coordination gap denotes the performance difference between the joint and modular solvers, quantified by the \gls{rpi} computed using $C_{\max}^{i}$ from the joint solver and $C_{\max}^{y}$ from the modular solver.

To quantify the availability of transportation resources relative to production demand, 
the resource scarcity factor ($\rho$) represents the ratio of the number of \glspl{agv} ($k$) to the total number of jobs ($n$):
\begin{equation}
    \rho = \frac{k}{n},
\end{equation}
where $\rho \in [\frac{1}{n}, 1]$. 
The boundary condition $\rho = 1$ denotes the unconstrained transport limit, where each job can theoretically be serviced by a dedicated vehicle, 
thereby minimizing resource contention. 

To evaluate the sensitivity of the coordination gap to relative scheduling task durations, 
the temporal-dominance index ($\tau^*$) characterizes the relative weight of logistics versus processing. 
Let $p_{raw}$ be the average processing and $t_{raw}$ the average transportation times. To ensure scale-invariance, 
these values are normalized to a unit interval $[0, 1]$ using the system's global time bounds $[T_{min}, T_{max}]$:
\begin{equation}
  p' = \frac{p_{raw} - T_{min}}{T_{max} - T_{min}}, \quad t' = \frac{t_{raw} - T_{min}}{T_{max} - T_{min}},
\end{equation}with $T_{min} = 1$ and $T_{max} = 100$. Subsequently, the normalized ratio $\phi \in [0, 1]$ is computed, 
representing the processing share of the total "normalized temporal budget":\begin{equation}
    \phi = \frac{p'}{p' + t'}.
\end{equation}Finally, the ratio $\phi$ is rescaled to the symmetric interval $[-1, 1]$ to obtain the dominance index:
\begin{equation}
    \tau^* = 2\phi - 1 = \frac{p' - t'}{p' + t'}.
\end{equation}
\subsection{Training Details}
This section details the instance generation procedure for the training phase. 
To facilitate learning across diverse settings, \gls{jsspt} instances of size $n \times m \times k$ are generated stochastically for the number of jobs $n$, machines $m$, and \glspl{agv} $k$.
Operation-processing times are sampled from a discrete uniform distribution $DU(1, 100)$, 
with transportation times between machines drawn from the same distribution. 
According to the instance size ($n \times m$), 
$k$ is sampled uniformly from $DU(3,n)$. 
The training protocol incorporates diverse instance sizes, 
which are sampled from a predetermined set to facilitate robust learning across different job and machine ratios 
$(6\times 6, 10\times10, 15\times10, 20\times5, 30\times10)$.
Together with a sampled number of \glspl{agv}, this configuration of $n\times m\times k$ is used to train the agents for one rollout. 

\subsection{Implementation Settings}
The hyperparameters for the training settings are summarized in Table~\ref{table_hyperparameters}. 
To prevent overfitting to small-scale instances, 
the rollout length is adjusted dynamically to ensure each batch consistently encompasses four completed episodes.
Given that rewards are exclusively provided at the end of an episode, 
a high discount factor $\gamma$ is employed to facilitate the propagation of reward signals across states. 
To facilitate reproducibility and provide the full hyperparameter configuration, the source code is available at \url{https://github.com/proto-lab-ro/jsspt-coordination-gap}.
\begin{table}[h]
\vspace{0.2cm}
\caption{Hyperparameter configuration}
\vspace{-0.2cm}
\label{table_hyperparameters}
\begin{center}
\begin{tabular}{ccc}
\toprule
\textbf{category} & \textbf{parameter} & \textbf{value} \\ \midrule
\multirow{7}{*}{training} & total frames & $4 \times 10^{6}$ \\
 & optimizer & Adam \\
 & learning rate & 0.0003 \\
 & learning rate schedule & linear decay \\
 & instances per rollout & 4 \\
 & number epochs & 1 \\ 
 & scaling factor $s$&5 \\ \midrule
\multirow{5}{*}{\gls{ppo} settings} & clipping $\epsilon$ & 0.2 \\
 & discount factor $\gamma$ & 0.999 \\
 & GAE parameter $\lambda$ & 1.0 \\
 & entropy coefficient & 0.01 \\
 & critic (value) coefficient & 0.5 \\ \bottomrule
 \end{tabular}
 \end{center}
\vspace{-0.4cm}
\end{table}

\subsection{Comparative Performance Evaluation and Sensitivity to Resource Scarcity}
In the first experiment, we compare the solvers against \gls{dr} baselines. For the evaluation, instances are sampled from the experimental configurations, summarized in Table~\ref{table_experiment_configs}. Processing durations for the instances are drawn from a discrete uniform distribution $DU(1, 100)$, 
consistent with the training distribution. 
To assess performance across varying levels of resource availability, 
the number of \glspl{agv} is parameterized using the resource scarcity factor $\rho \in \{0.2, 0.4, 0.6, 0.8, 1.0, 1.2\}$. For each configuration, 100 unique instances are processed by each solver to ensure statistical significance. 
Performance is evaluated by comparing the solvers against different combinations of \glspl{dr}. First, the \gls{rpi} is computed relative to the best-performing \gls{dr} combination for each individual instance. In subsequent evaluations, the best \glspl{dr} are excluded, and performance is assessed against the best combination from the remaining set of \glspl{dr}. Finally, three additional benchmarks are conducted in which solver performance is compared to fixed \gls{dr} combinations, ranging from the best to the third-best, as determined by the \gls{wr} aggregated over all configurations.
\begin{table}[b]
\vspace{-0.5cm}
\caption{Experiment configuration}
\vspace{-0.2cm}

\begin{center}
\begin{tabular}{cc}
\toprule
\textbf{instance: jobs $\times$ machines  }& \textbf{$\times$ \glspl{agv} }\\\midrule 
15$\times$10  & 18, 15, 12, 9, 6, 3 \\

10$\times$10   & 12, 10, 8, 6, 4, 2 \\

12$\times$12  & 14, 12, 10, 7, 5,2\\

14$\times$14  & 17, 14, 11, 8, 6, 3 \\

20$\times$5   & 24, 20, 16, 12, 8, 4 \\

5$\times$10   & 6, 5, 4, 3, 2, 1\\

15$\times$15 & 18, 15, 12, 9, 6, 3 \\

30$\times$10  &36, 30, 24, 18, 12, 6 \\
\bottomrule
\end{tabular}
\label{table_experiment_configs}
\end{center}
\vspace{-0.4cm}
\end{table}

\begin{figure*}[t]
    \vspace{1mm}
    \centering
    % Left Side: The Mean & 95% CI Plot
    \begin{subfigure}[t]{0.33\textwidth}
        \centering
        \begin{tikzpicture}

\begin{axis}[
    width=\linewidth,
    height=6.0cm,
    ylabel={mean \gls{rpi} (\%)},
    ylabel style={yshift=-4mm},
    xtick=data,
    xticklabels={vs all, vs w/o top 1, vs w/o top 2, vs w/o top 3},
    x tick label style={rotate=30, anchor=north east, font=\small},
    ymin=-7.2, ymax=6.2,
    grid=both,
    grid style={dashed, gray!20},
    ytick distance=2,
    extra y ticks={0},
    extra y tick style={grid style={solid, black}, tick label style={draw=none}},
    axis line style={black!80},
    enlarge x limits=0.015,
    legend cell align={left},
    legend style={
    %legend columns=7,
    %column sep=6pt,
    row sep=0pt,
    at={(0.00,1.0)},
    anchor=north west,
    %row sep=0pt,
    draw=none,
    fill=none,
    font=\fontsize{8}{10}\selectfont
    }
]
% =========================
% COUPLED (REFERENCE LINE)
% =========================
\addplot[
    thick,
    mark=square*,
    mark options={fill=darkgray},
    mark size=2pt,
    error bars/.cd,
    %legend image post style={scale=1.2}
    y dir=both,
    y explicit,
    error bar style={thick, darkgray}
] coordinates {
(0, -2.097978) +- (0, 0.210945)
    (1, -0.040609) +- (0, 0.210282)
    (2, 2.587876) +- (0, 0.208100)
    (3, 5.767222) +- (0, 0.250222)
};
\addlegendentry{joint: J0\_A0}

% =========================
% INDIVIDUAL DECUPLED LINES
% =========================

\addplot[
    name path=upper,
    thick,
    mark=*,
    mark size=2pt,
    mark options={fill=darkgray},
    error bars/.cd,
    y dir=both,
    y explicit
] coordinates {
(0, -5.122741) +- (0, 0.337796)
    (1, -2.954538) +- (0, 0.320522)
    (2, -0.191603) +- (0, 0.299402)
    (3, 3.070300) +- (0, 0.329914)
};
\addlegendentry{best: J2\_A1}

\addplot[
    name path=lower,
    thick,
    mark=triangle*,
    mark size=2pt,
    mark options={fill=darkgray},
    error bars/.cd,
    y dir=both,
    y explicit
] coordinates {
(0, -6.938536) +- (0, 0.292731)
    (1, -4.717996) +- (0, 0.269615)
    (2, -1.853149) +- (0, 0.227124)
    (3, 1.431664) +- (0, 0.279220)
};
\addlegendentry{worst: J3\_A7}

\addplot[
    fill=gray,
    fill opacity=0.25,
    draw=none
] fill between[
    of=upper and lower
];
\addlegendentry{Range top 20}

\end{axis}
\end{tikzpicture}
        \caption{\gls{rpi} performance across best-performing \gls{dr} combination per instance.}
        \label{fig:solver_rpi_95_best_and_total}

    \end{subfigure}
    \hfill
    \begin{subfigure}[t]{0.33\textwidth}
        \centering
        \begin{tikzpicture}

\begin{axis}[
    width=\textwidth,
    height=6cm,
    ylabel={mean \gls{rpi} (\%)},
    ylabel style={yshift=-4mm},
    xtick=data,
    xticklabels={vs top 1 \gls{dr},vs top 2 \gls{dr}, vs top 3 \gls{dr}},
    x tick label style={rotate=30, anchor=north east, font=\small},
    ymin=-7.2, ymax=6.2,
    grid=both,
    grid style={dashed, gray!20},
    ytick distance=2,
    extra y ticks={0},
    extra y tick style={grid style={solid, black}, tick label style={draw=none}},
    axis line style={black!80},
    enlarge x limits=0.015,
    legend cell align={left},
    legend style={
    %legend columns=7,
    %column sep=6pt,
    row sep=0pt,
    at={(0.00,1.0)},
    anchor=north west,
    %row sep=0pt,
    draw=none,
    fill=none,
    font=\fontsize{8}{10}\selectfont
    }
]
% =========================
% COUPLED (REFERENCE LINE)
% =========================
\addplot[
    thick,
    mark=square*,
    mark options={fill=darkgray},
    mark size=2pt,
    error bars/.cd,
    %legend image post style={scale=1.2}
    y dir=both,
    y explicit,
    error bar style={thick, darkgray}
] coordinates {

    (0, -0.978741) +- (0, 0.204395)
    (1, 3.409302) +- (0, 0.268701)
    (2, 5.273837) +- (0, 0.188853)
};

% =========================
% INDIVIDUAL DECUPLED LINES
% =========================

\addplot[
    name path=upper,
    thick,
    mark=*,
    mark size=2pt,
    mark options={fill=darkgray},
    error bars/.cd,
    y dir=both,
    y explicit
] coordinates {

    (0, -3.976379) +- (0, 0.333202)
    (1, 0.550692) +- (0, 0.368700)
    (2, 2.588650) +- (0, 0.275261)
};

\addplot[
    name path=lower,
    thick,
    mark=triangle*,
    mark size=2pt,
    mark options={fill=darkgray},
    error bars/.cd,
    y dir=both,
    y explicit
] coordinates {

    (0, -5.779009) +- (0, 0.290804)
    (1, -1.219757) +- (0, 0.344896)
    (2, 0.993425) +- (0, 0.192863)
};

\addplot[
    fill=gray,
    fill opacity=0.25,
    draw=none
] fill between[
    of=upper and lower
];

\end{axis}
\end{tikzpicture}
        \vspace{-4.5mm}
        \caption{\gls{rpi} performance across specific \gls{dr} combinations.}
        \label{fig:solver_rpi_95_best_and_totalv2}

    \end{subfigure}
    \hfill
    % Right Side: The Distribution Analysis (Dummy Plot)
    \begin{subfigure}[t]{0.32\textwidth}
        \centering
        \begin{tikzpicture}
\begin{axis}[
    width=\linewidth,
    height=6.4cm,
    xlabel={$\rho$},
    ylabel={mean \gls{rpi} (\%)},
    ylabel style={yshift=-4mm},
    grid=major,
    grid style={dashed, gray!30},
    legend pos=north east,
    axis line style={black!80},
    tick label style={font=\small},
    extra y ticks={0},
    extra y tick style={grid=none, tick pos=left, tick style={draw=none}},
    extra y tick labels={},
    ymajorgrids=true,
    % Ensure all x-ticks are visible
    xtick={0.2, 0.4, 0.6, 0.8, 1.0, 1.2},
    ytick={4,  3,  2,  1,  0},
]

% Zero line (baseline)
\addplot[black, dashed, forget plot] coordinates {(0.2,0) (1.2,0)};

% Confidence Interval (Fill area)
% Upper Bound: Mean + (CI_multiplier * Std)
\addplot[name path=upper, draw=none, forget plot] coordinates {
    (0.2, 0.982754) (0.4, 2.961436) (0.6, 2.814733) (0.8, 1.378462) (1.0, 0.640041) (1.2, -0.196467)
};
% Lower Bound: Mean - (CI_multiplier * Std)
\addplot[name path=lower, draw=none, forget plot] coordinates {
    (0.2, 2.344872) (0.4, 4.153728) (0.6, 4.081999) (0.8, 2.671114) (1.0, 1.918557) (1.2, 1.118787)
};

% Fill the area between upper and lower paths
\addplot[black!70, opacity=0.3, forget plot] fill between[of=upper and lower];

\addplot[
    black, 
    thick, 
    mark=*, 
    mark options={fill=black}
] coordinates {
    (0.2, 1.663813)
    (0.4, 3.557582)
    (0.6, 3.448366)
    (0.8, 2.024788)
    (1.0, 1.279299)
    (1.2, 0.461160)
};
%\addlegendentry{Mean RPI}

\end{axis}

\end{tikzpicture}
        \caption{\gls{rpi} of J0\_A0 vs. J2\_A1 relative to $\rho$ (mean $\pm$ 95\% \gls{ci}).}
        \label{fig:agv2job_bm}
 
    \end{subfigure}
    \caption{Performance analysis of joint and modular solvers. (a) and (b) Overall \gls{rpi} benchmark analysis of the joint solver and the top 20 modular solvers (best modular solver: J2\_A1, worst modular solver: J3\_A7) with top 1 \gls{dr}: \gls{mor} \& \gls{scpt}, top 2 \gls{dr}: \gls{fdd/mwr} \& \gls{scpt}, top 3 \gls{dr}: \gls{mor} \& \gls{scta}.
    (c) Behavior of \gls{rpi} relative to resource scarcity factor $\rho$ for joint and best modular solver.}
    \label{fig:full_solver_analysis}
    \vspace{-0.5cm}
\end{figure*}
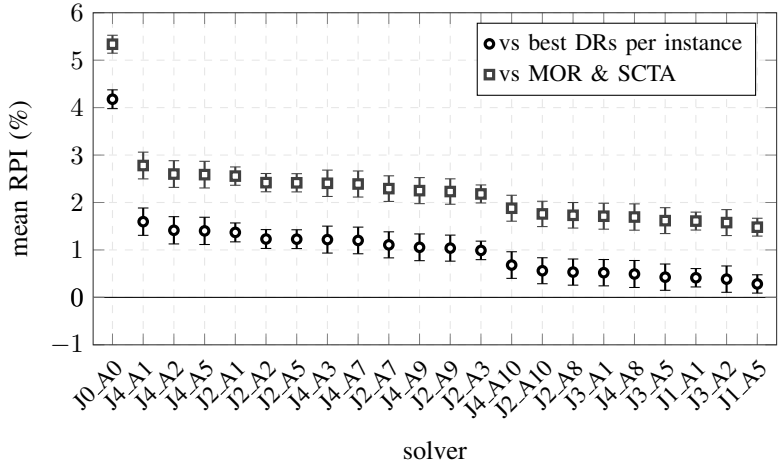
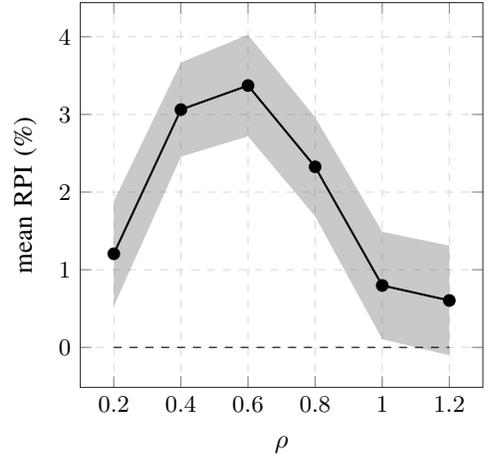

The results, illustrated in Figure~\ref{fig:solver_rpi_95_best_and_total} and~\ref{fig:solver_rpi_95_best_and_totalv2}, show the mean \gls{rpi} with its 95\% \gls{ci} across all evaluation instances for the joint solver and the top 20 modular solvers. 
Across benchmark settings, solver performance improves as top-performing \gls{dr} combinations are progressively removed from the comparison set. Once the three strongest combinations (out of 40) are excluded, all solvers surpass the remaining \gls{dr} baselines. While the best individual \gls{dr} outperforms all \gls{rl} solvers in direct comparison, they are superior to the remaining combinations. This indicates the superiority of the \gls{rl} solvers over the majority of \glspl{dr}.
Comparing the joint and modular solvers, the joint solver (J0\_A0) consistently achieves the strongest performance among learning-based methods, exceeding the best modular solver (J2\_A1) by approximately 3\% in \gls{rpi} across all baselines. In contrast, modular solvers exhibit performance variability across configurations, indicating sensitivity to the choice of heuristic pairings during independent training.

% FERTIG
The second experiment investigates the coordination gap by analyzing the \gls{rpi} as a function of $\rho$. 
Figure~\ref{fig:agv2job_bm} depicts the \gls{rpi} between the joint solver and the best-performing modular solver J2\_A1. 
Regarding the \gls{rpi} calculation in Equation~\ref{eq:rpi}, 
the makespans $C_{\max}^{i}$ and $C_{\max}^{y}$ correspond to the results of the J0\_A0 and J2\_A1 solvers, respectively. 
The results indicate that the competitive advantage of joint training is maximized in environments with balanced resources. 
For extreme values of $\rho$, representing either severe transport bottleneck ($\rho < 0.4$) or process-constrained regimes ($\rho > 0.8$), 
the performance gap narrows from 3.6\% to 0.5\%. 
These results suggest that under highly constrained settings, the marginal benefits of holistic coordination through joint training diminish, as system performance is dominated by a single bottleneck.
In contrast, balanced operational conditions enable joint training 
to effectively optimize the coupling between job scheduling and \gls{agv} allocation, thereby maximizing system efficiency.
Depending on available transport resources, 
these insights can guide the strategic selection of training modalities for \gls{jsspt} applications.

\subsection{Analysis of the Coordination Gap and Task Interdependence}
We design a grid experiment using range-based sampling to systematically explore the effect of relative temporal-dominance ($\tau^*$) and resource scarcity ($\rho$) on solver performance. The processing times of operations and the transportation times of \glspl{agv} are independently sampled 
from discrete consecutive intervals spanning 
$[1,100]$, defining 10 bins ($[1,10], [11,20], \dots ,[91,100]$). Each interval combination forms a grid cell (total 100 cells), and within each cell, 
processing and transport times are sampled uniformly from the corresponding bins, resulting in systematic coverage of the induced $\tau^*$ values. 
For each cell, 20 random instances are generated per configuration listed in Table~\ref{table_experiment_configs}, 
providing a statistically robust dataset across varying levels of resource scarcity and problem scales. 
The parameters $\rho$ and $\tau^*$ are integrated into a bivariate parameter space by grouping them according to their respective $\rho$ and $\tau^*$ values. This representation enables the characterization of the interdependence between transport resource allocation and operational duration dynamics. The resulting parameter space reveals four distinct operational regimes:
\begin{itemize}
    \item Underutilized transport ($\rho < 0.5$, $\tau^* > 0.0$): This regime forms no dominant bottleneck. It is characterized by a restricted number of \glspl{agv}, 
    paired with processing duration dominance. This prevents the formation of a dominant bottleneck.
    
    \item Process-constrained ($\rho > 0.5$, $\tau^* > 0.0$): In this regime, performance is governed by the underlying job scheduling 
    logic rather than transport constraints. With both abundant resource availability and high transport efficiency, 
    the system is primarily limited by machine processing. 
    
    \item Transport-constrained ($\rho < 0.5$, $\tau^* < 0.0$): This regime represents a critical transport bottleneck where 
    vehicle scarcity is exacerbated by transport duration dominance. The system performance is highly sensitive to the 
    transport scheduling.

    \item Resource-saturated ($\rho > 0.5$, $\tau^* < 0.0$): In contrast to the initial scenario, this regime features a dominance of transportation durations that is mitigated by abundant resource availability. This prevents 
    transport from becoming the primary system bottleneck.
\end{itemize}

\begin{figure}[tb]
    \centering
    \vspace{2pt}
    \pgfplotsset{compat=1.18}
\begin{tikzpicture}
\begin{axis}[
    xlabel={$\rho$},
    ylabel={$\tau^*$},
    % Custom Color Map (Red-Yellow-Green)
    colormap={rdylgn}{
        rgb255(0cm)=(255,255,255);  
        rgb255(10cm)=(97,97,97);
    },
    colorbar,
    colorbar style={title={mean \gls{rpi} rate (\%)},width=0.25cm,title style={xshift=-6mm}},
    point meta min=0,
    point meta max=7,
    % Grid Setup
    view={0}{90},
    enlargelimits=false,
    axis on top,
    % Label Formatting
    xtick={0,1,2,3,4,5},
    xticklabels={0.2, 0.4, 0.6, 0.8, 1.0, 1.2},
    ytick={0,2,4,6,8,10,12,14,16,18,20},
    yticklabels={-1.0, -0.8, -0.6, -0.4, -0.2, 0.0, 0.2, 0.4, 0.6, 0.8, 1.0},
    % Values inside cells
    nodes near coords,
    nodes near coords style={
        font=\fontsize{9}{5}\selectfont, % Very small font for dense data
        color=black,
        anchor=center,
        /pgf/number format/fixed,
        /pgf/number format/precision=1 % Show one decimal place
    },
    width=8cm, height=8cm % Adjusted for the 21x6 aspect ratio
]

\addplot[
    matrix plot*,
    mesh/cols=6, % Crucial: tells TikZ there are 6 data points per row
    point meta=explicit
] 
 table[meta=z] {
        x y z
        0 0 -0.2
        1 0 0.0
        2 0 0.9
        3 0 3.3
        4 0 6.2
        5 0 5.1
        
        0 1 0.2
        1 1 0.7
        2 1 1.2
        3 1 3.0
        4 1 6.2
        5 1 3.8
        
        0 2 0.1
        1 2 0.5
        2 2 1.6
        3 2 3.6
        4 2 5.1
        5 2 3.6
        
        0 3 0.0
        1 3 0.4
        2 3 1.3
        3 3 3.2
        4 3 5.6
        5 3 3.1
        
        0 4 0.2
        1 4 0.8
        2 4 1.8
        3 4 3.7
        4 4 5.0
        5 4 2.8
        
        0 5 0.6
        1 5 0.9
        2 5 2.2
        3 5 3.5
        4 5 3.9
        5 5 1.9
        
        0 6 0.6
        1 6 1.2
        2 6 2.4
        3 6 3.8
        4 6 3.7
        5 6 1.8
        
        0 7 0.9
        1 7 1.5
        2 7 2.7
        3 7 3.4
        4 7 2.8
        5 7 1.2
        
        0 8 1.1
        1 8 1.8
        2 8 2.9
        3 8 3.3
        4 8 2.2
        5 8 1.2
        
        0 9 1.3
        1 9 2.4
        2 9 3.3
        3 9 2.8
        4 9 1.9
        5 9 0.9
        
        0 10 1.8
        1 10 2.9
        2 10 3.4
        3 10 2.5
        4 10 1.5
        5 10 0.9
        
        0 11 2.0
        1 11 3.3
        2 11 3.4
        3 11 2.2
        4 11 1.4
        5 11 0.6
        
        0 12 2.4
        1 12 3.7
        2 12 3.1
        3 12 1.9
        4 12 1.2
        5 12 0.8
        
        0 13 2.8
        1 13 4.1
        2 13 2.6
        3 13 1.4
        4 13 0.6
        5 13 0.5
        
        0 14 4.0
        1 14 4.2
        2 14 2.4
        3 14 1.3
        4 14 1.0
        5 14 0.7
        
        0 15 4.7
        1 15 3.9
        2 15 1.9
        3 15 0.9
        4 15 0.5
        5 15 0.3
        
        0 16 5.9
        1 16 3.3
        2 16 1.6
        3 16 0.7
        4 16 0.9
        5 16 0.7
        
        0 17 6.5
        1 17 2.5
        2 17 1.4
        3 17 0.6
        4 17 0.6
        5 17 0.8
        
        0 18 6.0
        1 18 2.6
        2 18 1.2
        3 18 0.8
        4 18 0.6
        5 18 0.6
        
        0 19 6.7
        1 19 2.2
        2 19 1.3
        3 19 0.7
        4 19 1.1
        5 19 1.0
        
        0 20 4.8
        1 20 1.0
        2 20 0.9
        3 20 0.6
        4 20 0.6
        5 20 0.5
    };
\end{axis}
\end{tikzpicture}
    \vspace{-0.4cm}
    
    \caption{Mapping the coordination gap under coupling factors $\rho$ and $\tau^*$. 
        The heatmap displays the mean \gls{rpi} between the joint ($J0\_A0$) and modular ($J2\_A1$) solvers, with $C_{\max}^{i}$ from J0\_A0 and $C_{\max}^{y}$ from J2\_A1.
        }
    \label{fig:grid_joint_vs_modular}
    \vspace{-0.4cm}
\end{figure}
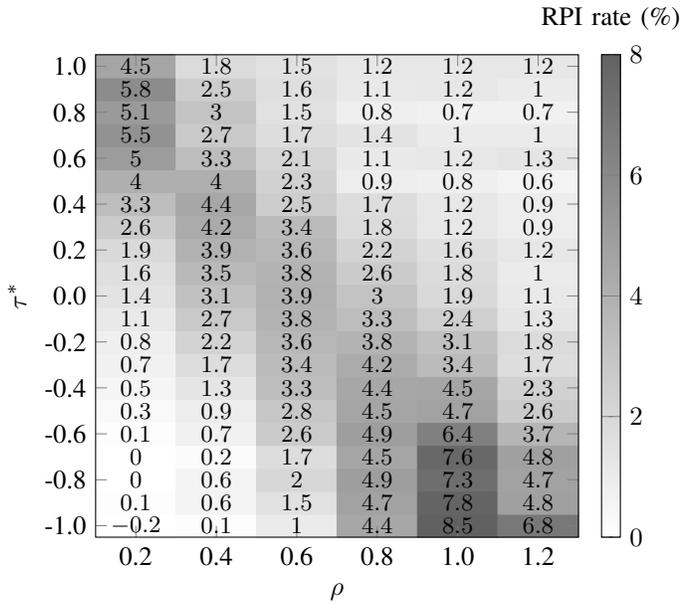

We utilize these regimes to analyze the interdependency of the two factors and the resulting performance delta between joint and modular training. Figure~\ref{fig:grid_joint_vs_modular} illustrates the \gls{rpi} of the joint solver ($J0\_A0$) in contrast to the best-performing modular solver ($J2\_A1$). 
The analysis reveals a distinct hierarchy of solver efficacy across the operational regimes: as illustrated, $J0\_A0$ demonstrates significant performance gains within the resource-saturated and underutilized transport regimes, with regime mean \glspl{rpi} of $3.0\%$ and $3.8\%$, respectively. 
In these regions, the absence of a dominant bottleneck allows the joint solver to optimize the schedule more effectively, with peak \gls{rpi} values reaching up to $6.7\%$. Conversely, in the process-constrained and transport-constrained bottleneck regimes, the \gls{rpi} decreases and solver performance equalizes. Within the transport-constrained regime, several cells show zero improvement, and one cell even yields a slightly negative value. Consequently, the mean \glspl{rpi} for these bottleneck regimes decrease to $1.1\%$ and $0.8\%$, respectively. 
The influence of the interplay between scheduling tasks is quantified by a hierarchical regression approach. \Gls{ols} regression models are fitted with the \gls{rpi} between the joint solver $J0\_A0$ 
and the best modular solver $J2\_A1$, as the dependent variable. The model variables are derived from the two coupling factors $\rho$ and $\tau^*$, to capture the balance and dominance of the tasks. 
First, the bottleneck dominance $BD$ is computed as:
\begin{equation}
    BD = |-\max(0, \tau^*) + (-\rho +1)|,
\end{equation} where the $\max$ term captures the dominance of the processing times, 
and the second term defines the \gls{agv} availability bottleneck region. 
Together they denote two distinct bottleneck regimes, 
where either the job or transport scheduling task is dominant. 
The balance metric $BM$ is then defined as $BM = (BD-1)^2$. 
For the influence of the bottleneck dominance, two variables are defined:
\begin{equation}
    JBN = \tau^* \times  \rho, \quad ABN = (\rho-1) \times \tau^*.
\end{equation}$JBN$ identifies the process-constrained regime with positive values and the resource-saturated regime with negative values. $ABN$ captures the underutilized transport and transport-constrained regimes with negative and positive values, respectively. In both cases, positive values highlight regimes where a single scheduling task creates a system bottleneck.
Negative values indicate balanced operational regimes, characterized by equally represented scheduling tasks.
The remaining regions of both variables exhibit values close to zero. For the following experiments, the features are $z$-normalized.

In the initial stage, the three features are individually analyzed by fitting an \gls{ols} regression model for each feature separately. 
The results indicate that all three features significantly influence the \gls{rpi}, yielding $R^2$ values of 0.43, 0.26, and 0.19 for $BM$, $ABN$, and $JBN$, respectively. 
The coefficient for $BM$ is positive (3.31), whereas $ABN$ (-3.00) and $JBN$ (-1.48) have negative coefficients. 
A multivariate model is fitted using $ABN$ and $JBN$ to determine the joint influence of bottleneck dominance. 
This model shows an $R^2$ value of 0.62 with negative coefficients (-3.94) for $ABN$ and (-2.09) $JBN$. 
Finally, a model with all three features is summarized in Table~\ref{table_ols_stats}. 
With an $F$-statistic of 71.34 ($p < 0.001$), 
the model confirms a statistically significant relationship. 
Multicollinearity is assessed using \gls{vif} values to ensure the reliability of the coefficient estimates. All \gls{vif} scores in Table~\ref{table_ols_coef} remain below five, indicating that moderate correlations exist, but critical multicollinearity is absent. The correlation between the features is:
\begin{itemize}
    \item $JBN$ and $BM$: -0.42
    \item $JBN$ and $ABN$: -0.26
    \item $ABN$ and $BM$: -0.43
\end{itemize}

The magnitudes of the coefficients are shown in Table~\ref{table_ols_coef}. 
While the full model achieves the highest $R^2$ (0.64), 
the marginal increase over the two-feature model is relatively modest. 
This suggests that the balance metric provides additional explanatory power, but the variance in \gls{rpi} is predominantly captured by the bottleneck dominance features. \begin{table}[b]
\vspace{-0.5cm}
\caption{\gls{ols} regression statistics}
\vspace{-0.2cm}
\label{table_ols_stats}
\begin{center}
\begin{tabular}{cc}
\toprule
\textbf{parameter}  & \textbf{value} \\ 
\midrule
$R^2$ & 0.64   \\
adj. $R^2$ & 0.63   \\
$F$-statistic & 71.34   \\
Prob (F-statistic)&$1.03\times 10^{-26}$ \\
observations & 126   \\
cond no. & 2.43   \\    

\bottomrule
\end{tabular}

\end{center}
\vspace{-0.4cm}
\end{table}

\begin{table}[t]
\setlength{\tabcolsep}{4pt}
\vspace{0.2cm}
\caption{\gls{ols} model coefficients}
\vspace{-0.2cm}
\label{table_ols_coef}
\begin{center}
\begin{tabular}{ccccccc}
\toprule
\textbf{variable} &\textbf{coef} &  \textbf{std error} & \bm{$t$} &  \bm{$P >|t|$} &  \textbf{95\% \gls{ci}} & \textbf{\gls{vif}} \\ 
\midrule
const & 2.16& 0.09& 24.70 & 0.0&  $[1.99, 2.34]$ & -\\
$BM$  &0.32& 0.12&  2.62& 0.01&   $[0.08,0.57]$& 1.97\\
$JBN$ & -0.80& 0.12&  -6.96& 0.0& $[-1.03, -0.58]$& 1.72\\
$ABN$ & -0.90& 0.12&  -7.76&  0.0& $[-1.13,-0.67]$& 1.75\\
\bottomrule
\end{tabular}
\end{center}
\vspace{-0.4cm}
\end{table} 

In all analyses, a trend of decreasing \gls{rpi} values is observed, corresponding to the increasing dominance of a single scheduling task, while regions with a balanced distribution of tasks exhibit higher \gls{rpi} values.
The negative coefficients for $JBN$ and $ABN$ lead to a decreasing \gls{rpi} in regions with scheduling task dominance and increase the \gls{rpi} in balanced regions. The positive coefficient of the balance metric yields an increase in \gls{rpi} with growing balance between tasks. 
These effects align with the observed behavior in the heatmap (Figure~\ref{fig:grid_joint_vs_modular}), in which the same trend is visible in the regimes. These insights indicate that the coordination gap is driven by the level of functional interdependence between scheduling tasks. The strong balance between transport and processing tasks enables joint training to effectively optimize the coupling between job scheduling and \gls{agv} allocation. In contrast, the presence of a dominant bottleneck diminishes the marginal benefits of holistic coordination. With a bottleneck, the system's performance is primarily constrained by a single scheduling task, thereby reducing the relative advantage of joint training over modular approaches.

\section{CONCLUSION}
This study demonstrates the influence of resource availability and temporal dynamics on the performance 
of joint versus modular training strategies in \gls{jsspt} tasks. 
Overall, joint training of job and \gls{agv} scheduling agents outperforms modular approaches. 
However, this advantage diminishes in bottleneck environments, where transport or processing constraints dominate and a hierarchical structure of scheduling tasks exists.
These findings have practical implications for the design of \gls{rl}-based scheduling systems in manufacturing environments. 
Depending on environmental conditions, decision-makers can select the most appropriate training modality to optimize performance and resource utilization.
This work provides a foundation for extending the \gls{marl} framework to industrial \gls{jsspt} scenarios. 
Future research will investigate flexible \gls{jsspt} with the inclusion of a machine-selection agent, as well as the integration of multiple objectives, including energy consumption, delay minimization, and cost efficiency.

\addtolength{\textheight}{-12cm}   % This command serves to balance the column lengths

%%%%%%%%%%%%%%%%%%%%%%%%%%%%%%%%%%%%%%%%%%%%%%%%%%%%%%%%%%%%%%%%%%%%%%%%%%%%%%%%

%%%%%%%%%%%%%%%%%%%%%%%%%%%%%%%%%%%%%%%%%%%%%%%%%%%%%%%%%%%%%%%%%%%%%%%%%%%%%%%%

%%%%%%%%%%%%%%%%%%%%%%%%%%%%%%%%%%%%%%%%%%%%%%%%%%%%%%%%%%%%%%%%%%%%%%%%%%%%%%%%

%%%%%%%%%%%%%%%%%%%%%%%%%%%%%%%%%%%%%%%%%%%%%%%%%%%%%%%%%%%%%%%%%%%%%%%%%%%%%%%%

\bibliographystyle{IEEEtran}
\bibliography{paper}

\end{document}